\documentclass[sigconf]{acmart}

\usepackage[utf8]{inputenc}

\usepackage{CJKutf8}
\usepackage{microtype}
\usepackage{algorithm}
\usepackage{booktabs} 
\usepackage{graphicx}
\usepackage{amsfonts}
\usepackage{amsmath}
 \usepackage{subfigure}
\usepackage[T1]{fontenc}
\usepackage{makecell}
\usepackage{multirow}
\usepackage{diagbox}

\usepackage{algpseudocode}
\DeclareMathOperator*{\argmin}{arg\,min}

\usepackage{bm}

\copyrightyear{2022} 
\acmYear{2022} 
\setcopyright{acmcopyright}\acmConference[SIGIR '22]{Proceedings of the 45th International ACM SIGIR Conference on Research and Development in Information Retrieval}{July 11--15, 2022}{Madrid, Spain}
\acmBooktitle{Proceedings of the 45th International ACM SIGIR Conference on Research and Development in Information Retrieval (SIGIR '22), July 11--15, 2022, Madrid, Spain}
\acmPrice{15.00}
\acmDOI{10.1145/3477495.3532071}
\acmISBN{978-1-4503-8732-3/22/07}

\begin{document}
\fancyhead{}
\title{Unifying Cross-lingual Summarization and Machine Translation with Compression Rate}

\author{Yu Bai$^{1,2,3}$, Heyan Huang$^{1,2,3}$, Kai Fan$^{\dagger}$,\\ Yang Gao$^{1,2,3}$, Yiming Zhu$^{1}$, Jiaao Zhan$^{1}$, Zewen Chi$^{1}$, and Boxing Chen}
\email{{yubai,hhy63,gyang,zym,jiaao\_zhan,czw}@bit.edu.cn}
\email{{interfk,chenboxing}@gmail.com}

\affiliation{%
  \institution{$^1$School of Computer Science, Beijing Institute of Technology, Beijing, China}
  \institution{$^2$Southeast Academy of Information Technology, Fujian, China}
  \institution{$^3$Beijing Engineering Research Center of High Volume Language Information Processing \\and Cloud Computing Applications, Beijing, China}
  \streetaddress{}
  \city{}
  \state{}
  \country{}
  \postcode{}
}

\renewcommand{\shortauthors}{Bai, et al.}

\begin{abstract}
  \renewcommand{\thefootnote}{\fnsymbol{footnote}}
  \footnotetext[2]{Corresponding author.}
  
\renewcommand{\thefootnote}{\fnsymbol{footnote}}
Cross-Lingual Summarization (CLS) is a task that extracts important information from a source document and summarizes it into a summary in another language. It is a challenging task that requires a system to understand, summarize, and translate at the same time, making it highly related to Monolingual Summarization (MS) and Machine Translation (MT). In practice, 
the training resources for Machine Translation are far more than that for cross-lingual and monolingual summarization. Thus incorporating the Machine Translation corpus into CLS would be beneficial for its performance. However, the present work only leverages a simple multi-task framework to bring Machine Translation in, lacking deeper exploration.

In this paper, we propose a novel task, Cross-lingual Summarization with Compression rate (CSC), to benefit Cross-Lingual Summarization by large-scale Machine Translation corpus. Through introducing \textit{compression rate}, the information ratio between the source and the target text, we regard the MT task as a special CLS task with a compression rate of $100\%$. Hence they can be trained as a unified task, sharing knowledge more effectively. However, a huge gap exists between the MT task and the CLS task, where samples with compression rates between $30\%$ and $90\%$ are extremely rare. Hence, to bridge these two tasks smoothly, we propose an effective data augmentation method to produce document-summary pairs with different compression rates. The proposed method not only improves the performance of the CLS task, but also provides controllability to generate summaries in desired lengths. Experiments demonstrate that our method outperforms various strong baselines in three cross-lingual summarization datasets.
We released our code and data at \url{https://github.com/ybai-nlp/CLS_CR}.

\end{abstract}

\ccsdesc[500]{Information systems~Summarization}
\ccsdesc[500]{Information systems~Multilingual and cross-lingual retrieval}


\keywords{Cross-lingual Summarization; Machine Translation; Compression Rate}


\maketitle
\section{Introduction}
Nowadays, acquiring information in a foreign language has become an important requirement for people of any nation. Millions of new documents are produced every day on the Internet. 
These documents are written in various languages, bringing more difficulties for people who are not familiar with other languages.
Hence, how to extract and understand the important information from documents in foreign languages is a meaningful and challenging research problem.
To achieve such a goal, Cross-lingual Summarization has been continuously explored by plenty of researchers~\citep{zhu2019ncls,zhu2020attend, cao2020jointly,ouyang2019robust}.

\begin{figure*}[t]
\centering
\includegraphics[width=1.99\columnwidth]{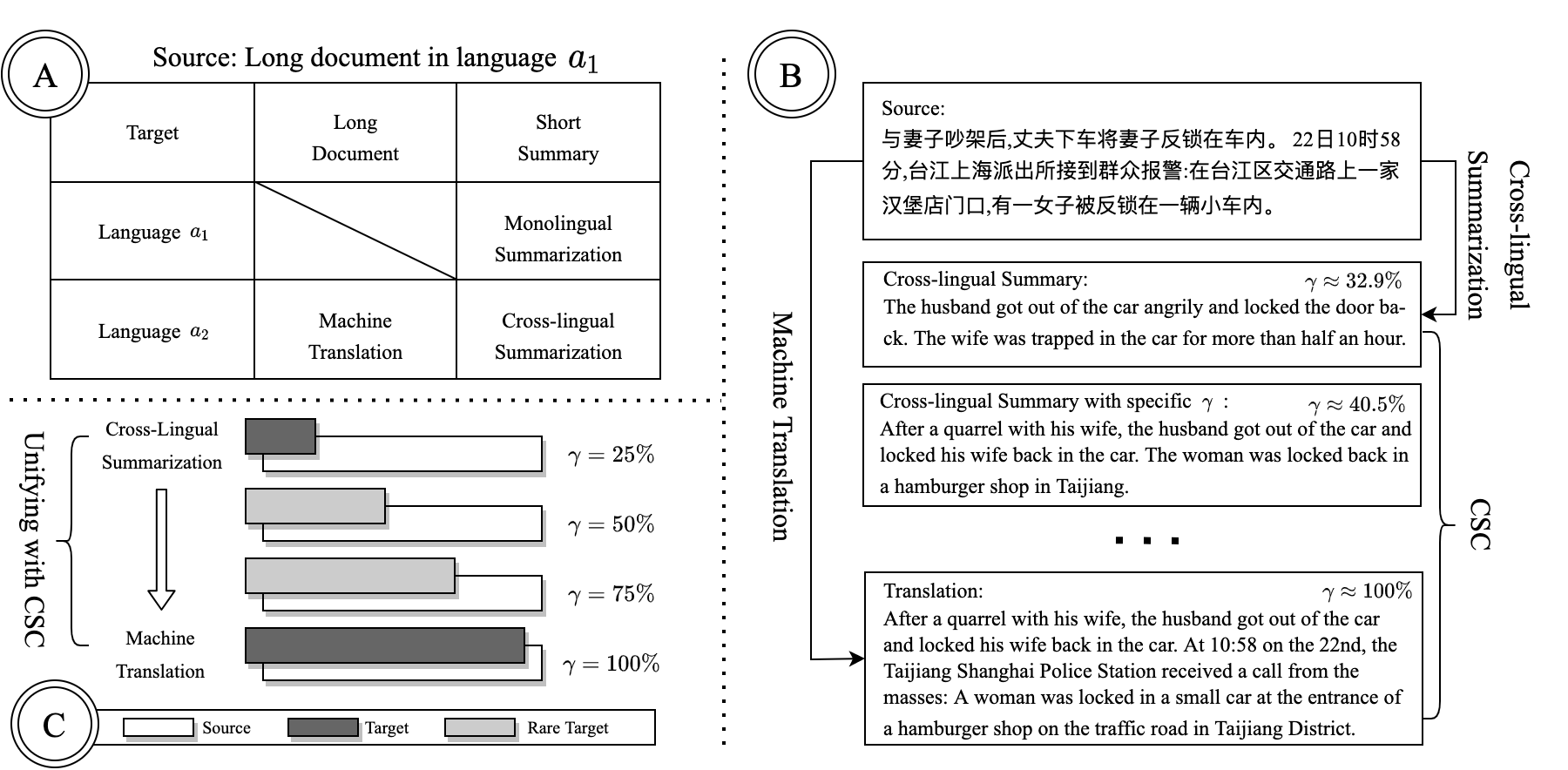} 
\caption{Overview: (A) shows the task definitions of Monolingual Summarization, Cross-lingual Summarization, and Machine Translation, where they produce different text output with the same source document in language $a_1$. (B) shows an example of different tasks including our proposed CSC, which unifies both CLS and MT to produce summaries in specific compression rates. Specifically, $\gamma$ represents the compression rate. (C) illustrates the relationship between CLS and MT. The area of the text square represents its text length. The lighter color of the target means that training samples with their $\gamma$ are rare.  }
\label{fig:cr}
\end{figure*}

Cross-lingual Summarization (CLS) aims at converting a document from one language to a summary in another language. 
It combines Machine Translation (MT) and Monolingual Summarization (MS), requiring the ability to both extract key ideas and translate them into the target language, illustrated in Figure~\ref{fig:cr}.
Exploring the relationship of these relevant tasks is important to improve the performance of CLS.
\citet{zhu2019ncls} apply the classic multitask framework of sequence-to-sequence model~\citep{luong2015multi}, using a unified encoder to share the knowledge between CLS and MT. 
\citet{takase2020multi} suggest using one single Transformer to learn the MT task, the MS task, and the CLS task, only distinguished by a special token.
As mentioned, CLS has a close relationship with the MT task.
However, these previous works simply treat MT as an independent and auxiliary task for CLS, lacking deeper exploration of their relationship. 
Hence, how to better leverage the huge MT corpus up to the hilt still remains a challenge. 

To achieve the mentioned goal, we thoroughly probe the relationship between MT and CLS. 
We observe that the MT task can be viewed as a special case of the CLS task via a concept in the summarization domain: \textit{compression rate}, which refers to the information ratio between the target summary and the original document in the summarization task~\citep{tas2007survey,babar2015improving,neto2002automatic,nomoto2001new}. 
If the target summary contains the exact same amount of information as the source document, the compression rate becomes 100\%. 
In a cross-lingual scenario, we assume that a CLS task with a compression rate of 100\%, is the MT task. 
In practice, the length ratio is commonly used to define the approximated compression rate in summarization task \citep{hahn2000challenges,yeh2005text}.

To make our hypothesis concrete, we unify MT and CLS by designing a novel task , Cross-lingual Summarization with Compression rate (CSC). CSC integrates the compression rate as an indicator variable into the sequence-to-sequence model to control how much information should be kept in the target summary. 
In our proposal, it is unnecessary to differentiate the CLS and MT task, in other words, the model merely learns the unified CSC task. 
During the unified model optimization, three purposes are expected, 1) the large parallel corpus of MT targets to improve the translation quality, 2) the summarization corpus enables the extraction of the key idea, 3) the compression rate controls the information amount of the output text. 

However, relying on the compression variable alone cannot directly achieve the integration of the MT and the CLS, because of the lack of diversity for compression rates in the training data.
The compression rate in the summarization task usually distributes around 25\%, leading to the data scarcity of compression rates between 30\% and 90\%. 
If the transition from CLS to MT is not smooth enough, the training of CSC will almost degenerate into regular multi-task learning. 
To bridge this gap, we propose a simple but effective method to generate augmented CLS samples with different compression rates. 
Given a well-annotated CLS data sample, we iteratively delete the less important language units (sentences and words) to shorten the source document, generating a document-summary pair with a larger compression rate. 
Intuitively, the compression rate transition from the CLS task to the MT task is illustrated in Figure~\ref{fig:cr}.

In order to integrate the continuous compression variable into the summarization model, e.g., Transformer \citep{vaswani2017attention}, we quantize the compression rates by grouping them into equally sized bins within $(0,1]$. 
We then introduce a set of embedding vectors to represent each bin. 
Taking as input the compression rate embedding, the quantitative signal enables the model to be aware of the information ratio of the sample. 
Moreover, adjusting the width of the bins can allow the trade-off between fine-grained controllability and overall generalization.

We conduct experiments in three CLS datasets: Zh2EnSum, En2Zh-Sum ~\citep{zhu2019ncls}, and WikiLingua Es-En~\citep{ladhak2020wikilingua}.  
Experimental results show that models trained with the
CSC task outperforms the strong baselines in various comparable conditions. 
Additionally, we show the powerful controllability of the proposed CSC models by generating summaries in different compression rates as desired, which further brings more practical value. 


Our main contributions can be described as follows:
\begin{enumerate}
\item We propose Cross-lingual Summarization with Compression rate (CSC), a novel task that unifies the MT task and the CLS task. A data augmentation method is proposed to generate CLS samples with different compression rates, bridging the two tasks.
\item Our proposed CSC is able to benefit CLS through large-scale MT corpora. We modify the traditional transformer architecture, using a specific compression rate embedding to model the ratio-based compression rate variable. 
\item Experiments demonstrate that models trained with CSC achieve better CLS performance than other tasks. Besides, CSC also shows great controllability by generating summaries in different desired lengths.
\end{enumerate}

\section{Related Work}
\subsection{Cross-lingual Summarization}
Cross-lingual summarization is a research topic that receives ascending attention from researchers. It combines the task of Machine Translation and Text Summarization, providing convenience for people to get information in foreign languages.

Traditional CLS systems produce cross-lingual summaries in a pipeline fashion, employing either first-translate-then-summarize or first-summarize-then-translate pipeline methods~\citep{wan2010cross, wan2011using,zhang2016abstractive}. 
Recently, end-to-end methods have been studied deeper. 
\citet{zhu2019ncls} propose the first large-scale CLS dataset, applying Transformer~\citep{vaswani2017attention} to this task. 
They further use a shared encoder and two separate decoders to conduct the MT task and the CLS task at the same time. 
Some researchers follow their work and propose various methods \citep{zhu2020attend, cao2020jointly, bai-etal-2021-cross}. 
More datasets are proposed recently \citep{ladhak2020wikilingua,perez-beltrachini-lapata-2021-models}, bringing new energy to this field.
However, the above methods pay less attention to how to fully utilize the enormous MT corpora. 
The multi-task model proposed by \citet{takase2020multi} is similar to our method, incorporating the MT task, the MS task, and the CLS task simultaneously. 
Meanwhile, with length information as input, they use a special length-ratio positional encoding to guide the model in generating summary of a specified length. 
However, they do not inspect the relationship between MT and CLS, lacking exploration to bridge the gap between the MT and the CLS.

\subsection{Fixed Length Text Summarization}
Fixed Length Summarization is a task that demands generating a target summary in a specific length. 
\citet{liu2018controlling} set the desired length as an input to the initial state of the decoder to control the output length.  \citet{takase2019positional} propose to use length-ratio position embedding to generate summaries in different lengths. \citet{makino2019global} use a length-constraint objective function to optimize the model. 

Our work also favors the fashion of length control. 
The proposed compression rate is used to manipulate the output length of the generated summary. 
However, our work is different from these works in two main aspects. 
First, the motivation of our method is to unify MT and CLS as one task to benefit the CLS task, not to generate fixed-length summaries. 
Second, by tuning the width of the compression bin, it allows for less fine-grained controllability to achieve better performance in a fixed compression rate without target length information.

\section{Background}
\subsection{Cross-lingual Summarization (CLS)}
Formally, we denote two different languages as $A$ and $B$. 
For the CLS task, a system converts a document $D^A = \{x_1^A, x_2^A,\dots, x_n^A\}$ in language~$A$ into a shorter summary $S^B = \{y_1^B, y_2^B \dots, y^B_m\}$ in language~$B$, where $x^A$ and $y^B$ represent tokens in language $A$ and $B$ while $n$ and $m$ are the lengths of $D^A$ and $S^B$, respectively.
Overall, the CLS task $f_{CLS}$ can be denoted as follows:
\begin{equation}
    S^B = f_{CLS}(D^A)
\end{equation}

Note that in a CLS dataset, 
it is assumed that the monolingual summary $S^A = \{x^A_1, x^A_2, \dots, x^A_{\hat{n}}\}$ is usually available, where the sequence length is denoted as $\hat{n}$. 
We will leverage these data to construct our augmented data in Section \ref{sec:aug}.

\subsection{Compression Rate}
Compression rate\footnote{Note it is different from the meaning in data compression, which is basically equal to the reciprocal of ours.} is a commonly-used concept in text summarization~\citep{tas2007survey,babar2015improving,neto2002automatic,nomoto2001new}. It indicates the information ratio between the target summary and the source document. 
Following the parlance in previous literature~\citep{hahn2000challenges,yeh2005text}, we define the compression rate $\gamma$ as the ratio of the text length between the summary and the document:
\begin{equation}
    \gamma = CR(D^A, S^B) = \frac{m}{n}
\end{equation}
This variable indicates the information compression degree of the summarization process. 
As $\gamma$ becomes larger, we expect the model to generate a less concise summary. 
It requires the model to be capable of identifying more important information and assembling a new sentence.

\subsection{Machine Translation}
Machine Translation (MT) aims at translating one sentence into a sentence in another language.
An MT system converts a source text $X^A = \{x'^A_1, x'^A_2, \dots, x'^A_{n'} \}$ in language~$A$ into its translated text $X^B = \{y'^B_1, y'^B_2, \dots, y'^B_{m'}\}$ in language~$B$. $x'^A$ and $y'^B$ are both tokens while $n'$ and $m'$ are the lengths of $X^A$ and $Y^B$.
The MT task $f_{MT}$ can be described as follows:
\begin{equation}
    X^B = f_{MT}(X^A)
\end{equation}

The scale of MT corpora is typically much larger than that of cross-lingual summarization. 
However, how to better leverage the MT corpora to enhance the CLS task remains a challenge. 

\begin{figure}[t]
\centering
\includegraphics[width=\columnwidth]{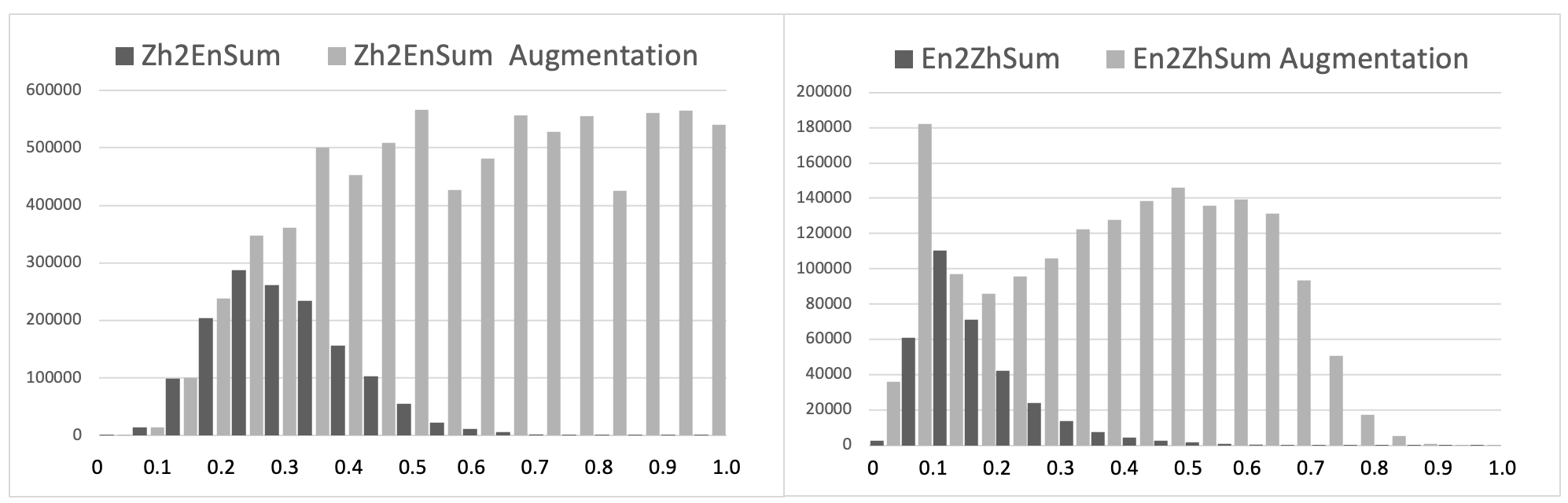} 
\caption{Compression rate statistics of Zh2EnSum and En2ZhSum training sets. The X-axis represents compression rate while the y-axis represents the number of training samples. Dark bars represent the original datasets while the light bars represent our augmented datasets. }
\label{fig:cr_statistics}
\end{figure}

\section{Methods}
To better utilize machine translation corpora, we propose a task that incorporates CLS and MT into one unified framework: Cross-Lingual Summarization with Compression Rate (CSC). 
Meanwhile, we propose to augment the current training data to better help the model produce summaries in different compression rates.

\subsection{Cross-Lingual Summarization with Compression Rate}

The main assumption of this work is that the MT task can be seen as a specific case of the CLS when the compression rate is $100\%$. 
To unify the two tasks, we propose Cross-lingual Summarization with Compression rate, producing a cross-lingual summary conditioned on both a source document $D^A$ and a compression rate $\gamma$:
\begin{equation}
    S^B = f_{CSC}(D^A, \gamma)
\end{equation}

With the compression rate variable $\gamma$, our model is able to view the CLS task and the MT task as a unified task, where $\gamma$ help the extraction of important content and the large-scale MT corpus can better ensure the quality (e.g., accuracy, fluency, and etc.) of the final summary.

However, directly training a CSC model on the existing MT and CLS corpora is nearly the same as regular multitask learning. 
Because the training pair with $\gamma = 30\% \to 90\%$ is rare, it leads to a huge gap between MT ($\gamma  \approx 100\% $) and CLS ($\gamma \approx 25\%$), shown in Figure~\ref{fig:cr_statistics}. 
To circumvent this problem, we design a data augmentation method to close it.

\subsection{Data Augmentation}
\label{sec:aug}
To bridge the Summarization tasks and Machine Translation tasks smoothly, we have to construct summarization samples with different compression rates. 
Specifically, the augmented summarization dataset should fulfill the following conditions:

\begin{itemize}
    \item  \textbf{Informative} The target summary should contain the important information of the source document, enabling the model to learn the summarization task.
    \item \textbf{Fluency} The target summary should be fluent and readable for the model to acquire the text generation ability.
    \item \textbf{Uniform} The compression rates should be approximately uniformly distributed within the interval $(0,1]$ to achieve the smooth transition from MT task to CLS task.
\end{itemize}

\begin{algorithm}[t]
\caption{Our proposed data augment method.}
\label{alg:augment}
\begin{algorithmic}[1]
\Require $D^A = \{s_1, s_2,\dots, s_l\}$, $S^A$, $S^B$, $\gamma$, $\hat{\gamma}$
\Ensure $\hat{\gamma} > \gamma$

\State $\hat{D^A} := D^A$ 

\State $\hat{s_i} := ROUGE\left(s_i, S^A \right)$,  $\forall i \in \{1,2,\dots, l\}$  
\State $k := \argmin_i \{\hat{s_i}\}_l$

\While {$CR(\hat{D^A}\setminus s_k, S^B)  < \hat{\gamma}$}
    \State $\hat{D^A} := \hat{D^A} \setminus s_k$
    \State $\hat{s_i} := rouge\left(s_i, S^A \right)$,  $\forall s_i \in \hat{D^A}$  
    \State $k := \argmin_i \{\hat{s_i}\}_n$
\EndWhile

\While {$CR(\hat{D^A}, S^B)  < \hat{\gamma}$}
    \State $w_i := $ a random word in $s_k$ but not in $S^A$
    \State $\hat{D^A} := \hat{D^A} \setminus w_i$
\EndWhile
\end{algorithmic}
\end{algorithm}
To achieve these goals, we propose a simple but effective data augmentation method. 
It is based on a simple fact: the more similar a sentence in the document is to the summary, the more important it is. 
This leads to an intuitive idea: we can keep the target summary unchanged, and gradually delete the less important sentences and words in the document to obtain the relatively larger compression rates. Note that samples with $\gamma$ less than the original CLS sample are not necessary to be augmented since our purpose is to bridge CLS and MT.

\begin{figure}[t]
\centering
\includegraphics[width=\columnwidth]{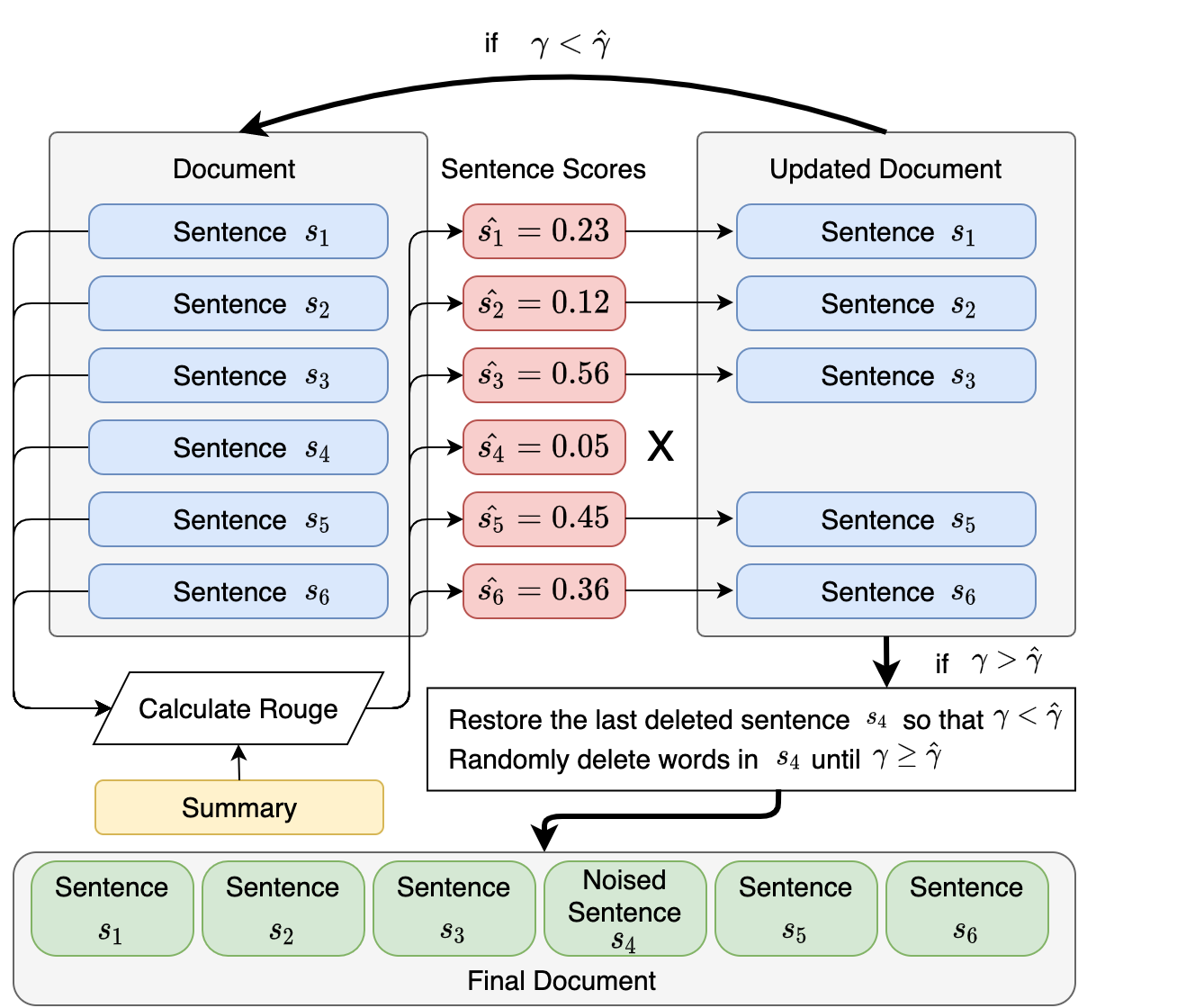} 
\caption{An illustration of our proposed compression rate based data augmentation method.}
\label{fig:augment}
\end{figure}

We leverage the monolingual summaries in CLS datasets to identify the importance of each sentence in the document. 
We calculate the ROUGE score \citep{lin2004rouge} between $S^A$ and each sentence $s_i$ in $D^A$. 
Then, we iteratively delete the least important sentence in $D^A$ to construct $\hat{D^A}$. 
The delete process continues until the compression rate of $\hat{D^A}$ and $S^B$ increases and reaches a desired $\hat{\gamma}$. 
However, because the sentence-level deletion is too coarse, the compression rate of the obtained document-summary pair could deviate from  $\hat{\gamma}$ a lot. 
So we restore the last deleted sentence such that the compression rate is still lower than $\hat{\gamma}$. 
Words in this sentence but not in the monolingual summary $S^A$ will be deleted randomly until a partially noisy document $\hat{D^A}$ with $\hat{\gamma}$ is acquired. 
Details of this process can be seen in Algorithm~\ref{alg:augment} and Figure~\ref{fig:augment}.

Our source-deleted augmentation method would possibly influence the fluency of the source document. However, it has been pointed out that the noise in source documents leads to a more robust CLS system~\citep{ouyang2019robust}.

We augment each sample with a sequence of ascending $\hat{\gamma}$s larger than its original $\gamma$. 
Concretely, the construction of the new compression rates can be characterized as injecting random perturbations to an arithmetic progression:
\begin{equation}
\small
    \{\hat{\gamma} \leq 1 | \hat{\gamma}= \gamma + (i + \mathcal{U}(0,1)) * 0.1 , i = 1,2,...,10\}
\end{equation}

The resulting augmented statistics of two CLS datasets are shown in Figure~\ref{fig:cr_statistics}\footnote{The data augmentation process is only performed on the training set.}.
Note that En2ZhSum is a more extractive dataset, where summaries tend to be extracted directly from the source document. Hence many words in the source document would appear in the target summary, leading to samples with higher $\gamma$ (around $90\%$) hard to constructed. 

\subsection{Model Architecture}

Transformer \citep{vaswani2017attention} is a widely-used architecture in Natural Language Processing. 
It consists of multiple stacked encoder and decoder Transformer layers. 
The encoder and decoder layers both have a self-attention block and a feed-forward block. 
Besides, the decoder layer also has an additional encoder-decoder attention block to acquire information from the source side. 
The multi-head attention is applied to all the attention modules.
Transformer has been applied to Cross-lingual Summarization in previous work \citep{zhu2019ncls, xu2020mixed}. 
In this paper, we introduce a novel embedding-based method to incorporate the compression rate into the Transformer model.

\begin{figure}[t]
\centering
\includegraphics[width=\columnwidth]{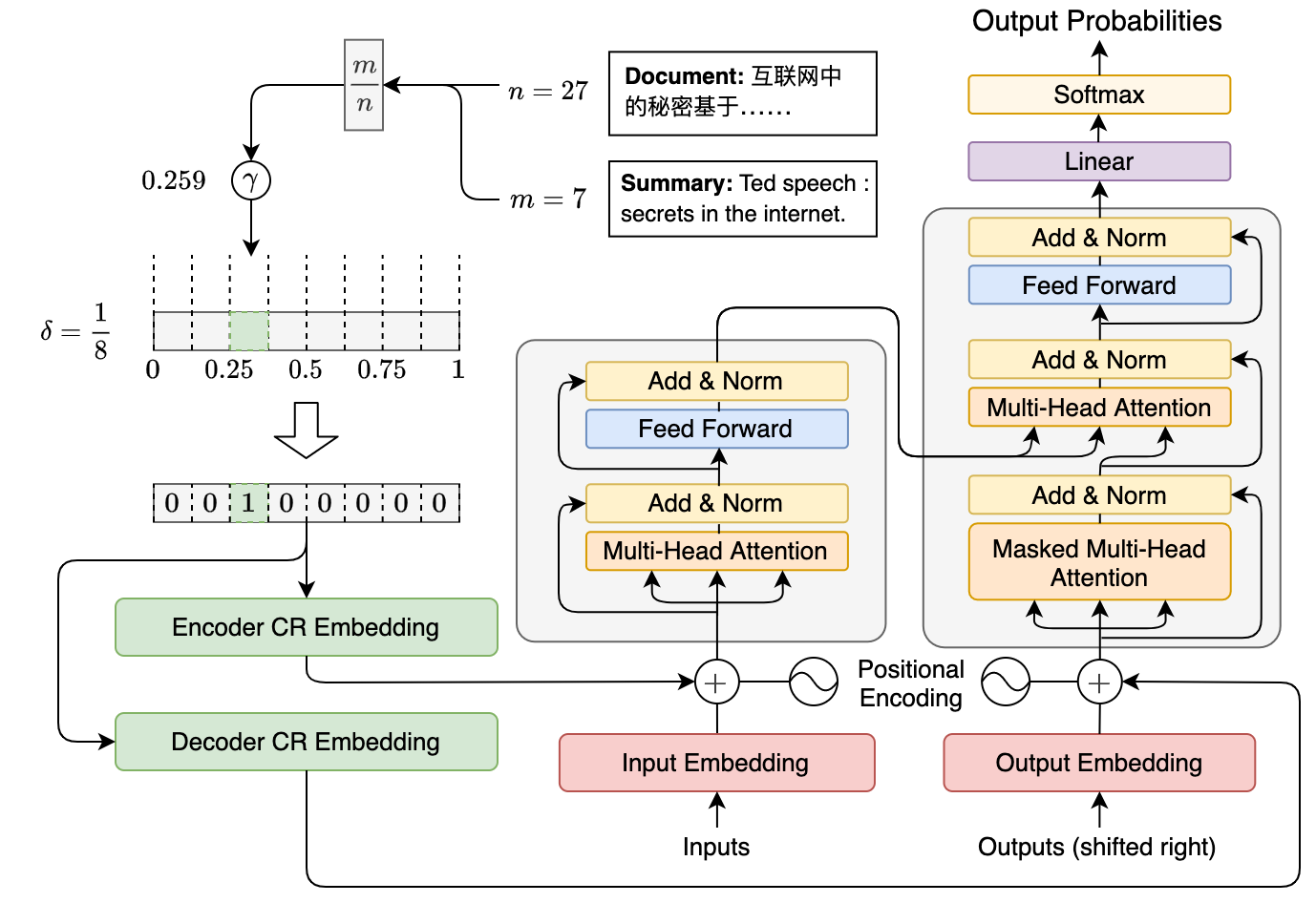} 
\caption{Model architecture of the modified Transformer incorporating compression rate, where $\gamma$ means the compresssion rate and $\sigma$ represents the bin size.}
\label{fig:model}
\end{figure}

Using specific embedding to control the discrete output length of the sequence-to-sequence model has been proved to be effective~\citep{fan2018controllable, kikuchi2016controlling}. However, for a ratio-based variable like compression rate, no such trials have been conducted to our knowledge. Considering the huge essential difference between the output length variable and the compression rate variable, we have to overcome several critical issues: 1) How to learn a continuous variable $\gamma$ for the Transformer model? 2) How to achieve a fairer comparison between a CSC model and a traditional CLS model?

In order to take as input the continuous compression rate $\gamma$, we apply the quantization trick to embed it into a numerical vector. 
We first divide $(0,1]$ into equally sized bins with the width $\delta$. 
Each $\gamma$ is grouped into the bin if it falls into the corresponding interval. 
Therefore, we can use a set of learned embedding vectors to represent each bin. 
The vector is added to the embedding of each token to make the model aware of the length of the target summary, constraining the compression rate in a specific range. 
Note that the token embedding of both the encoder and the decoder is modified. 
Details are illustrated in Figure~\ref{fig:model}.

Occasionally the compression rate of a few MT examples is $\geq 100\%$, but we simply clip $\gamma = 100\%$ for convenience. 
During training, the actual $\gamma$ of the training pair can be fed to the model. 
However, during inference, we cannot derive the exact compression rate from the source alone. 
As a compromise, we discuss two different settings, \textbf{fixed} and \textbf{oracle}, to evaluate our models from different aspects in Section~\ref{sec:baselines}. 
Moreover, we find that the selection of \textbf{bin width} $\delta$ is crucial, and will analyze the details in Section \ref{sec:bin}.


\section{Experimental Settings}
\subsection{Datasets}
We conduct our experiments on 3 CLS datasets: Zh2EnSum, En2Zh-Sum~\citep{zhu2019ncls}, and WikiLingua~\citep{ladhak2020wikilingua}.  
The Zh2EnSum dataset comes from the Chinese summarization dataset LCSTS~\citep{hu2015lcsts} using a round-trip strategy. 
The En2ZhSum dataset is converted from the MSMO dataset and the CNN/DM dataset constructed by the same method.
All the training samples contain a source document, a monolingual summary, and a cross-lingual summary. We apply the WMT17 Zh2En dataset for the Chinese-to-English translation data. 
The Es-En dataset in WikiLingua is constructed from WikiHow. 
Since WikiLingua contains documents and summaries in parallel languages, we directly leverage the document pairs and the summary pairs as MT training data. 
Detailed statistics are shown in Table \ref{tab:dataset}.

Specifically, we find repetition problems in the Zh2EnSum dataset. These problems appear in two aspects. First, the training set has many repetitions itself, we preprocess the dataset to eliminate these repetitions, and hence the training set contains 1,465,196 samples, which is smaller than the original dataset. Second, the test set contains samples appearing in the training set, which makes the evaluation of the test set less convincing. 
688 samples in the training set have repetitions and some of them appeared more than once.  Detailed statistics are shown in Table \ref{tab:repetition}.
We, therefore, divide the test set into two test sets Zh-En$_{\rm repeat}$ and Zh-En$_{\rm norepeat}$, containing 688 and 2,312 samples, to better evaluate our models. In order to compare with previous pipeline models, we also report the result of the full dataset Zh-En$_{\rm full}$.  We will release these sets to the public in the future.

\begin{table}[tb]
\caption{Statistics of the repetition problem.}
  \centering
  \resizebox{0.75\linewidth}{!}{
    \begin{tabular}{lrrrrrr}
    \toprule
    Repetition Times & 1 & 2 & 3 & 4 & 5 & $>$5 \\
    \midrule
    Num. of Samples & 390 & 122 & 44 & 35 & 26 & 71 \\

    \bottomrule
    \end{tabular}
    }

    \label{tab:repetition}
\end{table}

\begin{table}[t]
\caption{Statistics of all the datasets. Augmented means the augmented dataset of each CLS direction.}
  \centering
  \resizebox{0.99\linewidth}{!}{
    \begin{tabular}{lrrrrrr}
    \toprule
    Dataset & Train & Valid & Test & MT Train & MT Valid & Augmented \\
    \midrule
    Zh-En & 1,693,713 & 3,000 & 3,000 & 20,616,495 & 2,002 & 16,188,500 \\
    En-Zh & 364,687 & 3,000 & 3,000 & 20,616,495 & 2,002 & 1,709,753 \\
    Es-En & 81,514 & 9,057 & 22,643 & 163,028 & 18,114 & 423,006  \\

    \bottomrule
    \end{tabular}
    }

    \label{tab:dataset}
\end{table}

\begin{table*}[t]
    \caption{F1 scores of ROUGE in Zh2EnSum and En2ZhSum dataset. R-1, R-2, and R-L represents ROUGE-1, ROUGE-2, and ROUGE-L, respectively. The best results are bold while the least overfitting results are underlined.}
  \centering
  \resizebox{0.99\linewidth}{!}{
    \begin{tabular}{lcrrrrrrrrrrrrr}
    \toprule
    \multirow{2}*{Models} & \multirow{2}*{Parameters}  & \multicolumn{3}{c}{\textbf{Zh-En$_{\rm no repeat}$}} & \multicolumn{3}{c}{\textbf{Zh-En}$_{\rm repeat}$} & \multicolumn{3}{c}{\textbf{Zh-En}$_{\rm full}$}  & \multicolumn{3}{c}{\textbf{En-Zh}} 
    \\
    \cmidrule(r{4pt}){3-5} \cmidrule(r{4pt}){6-8} \cmidrule(r{4pt}){9-11} \cmidrule(r{4pt}){12-14}   
    ~ & &  R-1 & R-2 & R-L &  R-1 & R-2 & R-L &  R-1 & R-2 & R-L &  R-1 & R-2 & R-L \\
    \midrule
     NCLS & $\approx$300M & 34.95 & 15.99 & 30.21 & 49.38 & 31.51 & 44.98  & 39.34 & 20.66 & 34.77 & 42.34 & 22.80  & 28.56  \\
     TETran & \multirow{2}*{-} & - & - & - & - & - & - &23.09& 7.33 &18.74 & 26.15 & 10.60 & 23.24 \\
     TLTran & & - & - & - & - & - & - &  33.92& 15.81 &29.86 & 30.22 & 12.20 & 27.04  \\
     \midrule

     NCLS+AG(SA) & \multirow{4}*{$\approx$300M} & 35.29 & 16.65 & 30.48 & 48.23 & 30.39 & 43.82 & \textbf{39.68} & \textbf{20.94} & \textbf{34.98} & 43.01 & 22.71 & 28.29  \\
     NCLS+MT(SA) & ~ & 36.08 & 17.24 & 31.22 & 44.29 & \underline{25.37} & \underline{39.27} & 38.50 & 19.48 & 33.52 & 42.43 & 22.64 & 28.23 \\
     CSC$_{\rm base}$(fixed, $\delta=0.2$) & & 35.75 & 17.08 & 31.02 & \underline{43.78} & 25.42 & 39.39 & 37.94 & 19.28 & 33.20 & 42.35 & 22.89 & 28.63 \\
     CSC$_{\rm Multitask}$(SA) & & \textbf{36.10} & \textbf{17.30} & \textbf{31.30} & 45.51 & 26.98 & 40.70 & 38.60 & 19.84 & 33.85 & \textbf{43.12} & \textbf{22.96} & \textbf{28.50} \\
     \midrule
     NCLS+MT(SE) &  \multirow{3}*{$\approx$450M} & 35.66 & 17.04 & 30.99 & 44.71 & 26.45 &40.27  & 38.25 & 19.61 &  33.58 & \textbf{43.24} & 23.10 & 28.66 \\
     NCLS+MT(SD) &  & 35.09 & 16.22 & 30.18 & 46.23 & 26.83 & 41.23 &  38.83 & 19.66 &  33.81 & 43.05 & 22.75 & 28.26 \\
     CSC$_{768}$(fixed, $\delta=0.2$) &  & \textbf{36.13} & \textbf{17.33} &\textbf{ 31.22} & 46.53 & 28.41 & 42.06 & \textbf{38.96} & \textbf{20.29} & \textbf{34.11} & 42.97 & \textbf{23.28} & \textbf{28.91} \\
     \midrule
     CSC$_{\rm Multitask}$(SE) &\multirow{2}*{$\approx$600M} & 35.56 & 16.94 & 30.80 & 44.31 & 26.39 & 39.95 & 38.26 & 19.61 & 33.56 & 42.97 & 22.96 & 28.42 \\
     CSC$_{\rm Multitask}$(SD) & & 35.10 & 16.09 & 30.25 & 46.13 & 27.39 & 40.77 & 39.22 & 20.10 & 34.29 & 42.97 & 22.79 & 28.40 \\
     \midrule
     CSC$_{\rm base}$(oracle, $\delta=0.2$) & $\approx$300M & 36.76 & 17.76 & 31.98 & 45.26 & 26.35 & 40.79 &  39.12 & 20.18 & 34.37 & 43.65 & 23.26 & 28.82 \\
     CSC$_{768}$(oracle, $\delta=0.2$) & $\approx$450M & \textbf{37.06} & \textbf{18.03} & \textbf{32.21} & 47.84 & 29.79 & 43.44 & \textbf{40.30} & \textbf{21.43} & \textbf{35.46} & \textbf{44.03} & \textbf{23.60} & \textbf{29.18} \\

    \bottomrule
    \end{tabular}
    }

    \label{tab:ROUGE&BERTScore}
\end{table*}

\subsection{Baselines and Variations of CSC}
\label{sec:baselines}
We compare our proposed models with the following baselines:

\textbf{TETran} the pipeline model which translates the document and summarizes it~\citep{zhu2019ncls}.

\textbf{TLTran} the pipeline model which first summarizes a document into a summary and then translates the summary~\citep{zhu2019ncls}.

\textbf{NCLS} the vanilla baseline model which uses the Transformer to directly conduct CLS without any extra data.

\textbf{NCLS+MT} the Transformer-based multi-task framework which conducts CLS and MT simultaneously. To investigate its full potential, we experiment with its several variations:

\textbf{$\bullet$ Share Decoder (SD)}: the decoder is shared for both tasks, sharing the knowledge of how to generate text in the target language.

\textbf{$\bullet$ Share Encoder (SE)}:  the encoder is shared for both tasks, sharing the knowledge of how to get a better representation of source language.

\textbf{$\bullet$ Share All (SA)}: all the parameters are shared for these tasks. A task-specific token is before the text to distinguish different tasks.

\textbf{NCLS+AG (SA)} the Transformer-based multitask framework with both the CLS data and the augmented CLS data. This baseline is to ablate the effectiveness of the augmentation method.




Since models assisted by Monolingual Summarization (MS) corpora are empirically worse than those assisted by MT corpora. 
We omitted the baseline model NCLS+MS~\citep{zhu2019ncls}, which uses another separate decoder conducting MS task.

All the models in our experiments can handle both the Machine Translation task and the Cross-lingual Summarization task. 
However, models sharing partial parameters possess more parameters than those which share all parameters among tasks. 
Details can be seen in Table \ref{tab:ROUGE&BERTScore}. 
Hence, to achieve a fairer comparison, we set up two variations of our model for En2ZhSum and Zh2EnSum:

\textbf{$\bullet$ \bm{${\rm CSC_{base}}$}}: the base model of which the hidden dimension is 512. It has far fewer parameters than the above multitask models, leading to potentially unfair comparison.

\textbf{$\bullet$ \bm{${\rm CSC_{768}}$}}: the enhanced model with a hidden dimension of 768. It owns a similar capacity as multitask models.


For these two variations, we set $\delta=0.2$ as it can generate better summaries while maintaining controllability to a certain extent. The choice of $\delta$ is discussed in Section~\ref{sec:bin}. Also, as mentioned, two different inference modes are tested for these models. 
One is the \textit{fixed} compression rate mode, where a fixed $\gamma$ is fed into the model. 
The other is to test the model with the true (or desired) $\gamma$ of each sample, named as \textit{oracle}. 
It can verify the performance of our model when we have controllable signals during inference.
Selected by the validation performance, the second bin $[0.2, 0.4)$ is used as the fixed $\gamma$ at the inference stage for Zh2EnSum and Es-En while $(0, 0.2)$ is used for En2ZhSum.

To verify the effect of our proposed data augmentation with MT corpora, we also set another variation of our model:  \bm{${\rm CSC_{Multitask}}$}. 
Using augmented data as part of the training data, this variation treats CLS, MT, and the Augmented CLS as multiple tasks. 
We also conduct experiments with its SD, SE, and SA.

We show that our model outperforms all of the baseline models in all the comparable conditions. 
Details can be seen in Section~\ref{sec:experiment}.

\subsection{Implementation Details}
\label{sec:implementation}
We use Fairseq toolkit~\citep{ott2019fairseq} to implement all of our models with the dictionary of mBART~\citep{liu2020multilingual}, including 200,027 tokens. 
We use sentencepiece~\citep{kudo-richardson-2018-sentencepiece} to tokenize all the texts. Besides,
the compression rate is calculated by the number of tokenized tokens, which is intuitive since the model is only aware of them.
For the model architecture of En2ZhSum and Zh2EnSum, we follow the base Transformer settings proposed by \citet{vaswani2017attention}. 
All the Transformer encoders and decoders used in the model contain 6 layers. 
The size of feed-forward layers is 2,048. 
The hidden size for the attention module is either 512 or 768 depending on the model variation. 
Each attention layer contains 8 different attention heads. A task indicator is put before source text and target text for all the multitask models for them to distinguish different tasks. 
The parameter statistics are shown in Table~\ref{tab:ROUGE&BERTScore}. For the WikiLingua dataset, we use mBART to initialize all of our models. For multitask models with multiple encoders or decoders, we use the encoder or decoder of mBART to initialize them all. 

The Adam optimizer is used to train all the models with a learning rate of 5e-4 and 5000 warm-up steps.
We use a dropout of 0.2 for feed-forward layers and 0.1 for attention layers.
We truncate the source document to the length of 1024.
All the models are trained with 8 Tesla V100 32G GPUs. 
The batch size is 2048 tokens and the model parameters are updated every 16 batches. 
We apply the early stop strategy when the validation loss no longer improves for two checkpoints and choose the checkpoint with the best validation loss to conduct the evaluation.
During inference, we use beam search with a beam size of 5. Meanwhile, 3-gram blocking is applied to avoid repetition problems. 
All the hyper-parameters are tuned using the perplexity and validation loss metric on the validation set.
For all CSC models, the data among all the different tasks are fed into the model uniformly.

For mBART initialized models, the learning rate is set as 5e-5 and 5000 warm-up steps. Other settings are as same as the base models. Note that for Share All models of mBART there are no extra tokens in mBART dictionary for us to distinguish different tasks. Hence we use a similar way to achieve it: adding a trainable task-specific embedding to all the tokens of the encoder and decoder. 

\begin{table}[t]
    \caption{F1 scores of ROUGE in WikiLingua Es-En dataset. R-1, R-2, and R-L represents ROUGE-1, ROUGE-2, and ROUGE-L, respectively.}
  \centering
  \resizebox{0.85\linewidth}{!}{
    \begin{tabular}{lrrr}
    \toprule
    Models & R-1 & R-2 & R-L \\
    \midrule
    TETran & 37.16 & 14.25 & 31.04 \\
    TLTran & 36.03 & 13.02 & 29.86 \\
     NCLS(mBART) & 38.79 & 15.24 & 31.50 \\
     \midrule
     NCLS+AG(SA) & 38.76 & 15.02 & 31.15  \\
     NCLS+MT(SA) & 39.05 & 15.09 & 31.40  \\
     NCLS+MT(SE) & 39.17 & 15.22 &  31.42     \\
     NCLS+MT(SD) & 39.11 & 14.96 &  31.17  \\
     \midrule
     CSC$_{\rm Multitask}$(SA) & 39.24 & 15.25 & 31.20 \\
     CSC$_{\rm Multitask}$(SE) & 39.25 & 15.35 & \textbf{31.56} \\
     CSC$_{\rm Multitask}$(SD) & 38.91 & 15.10 & 31.26  \\
     \midrule
     CSC(fixed, $\delta = 0.2$) & \textbf{39.38} & \textbf{15.43} & 31.01 \\
     CSC(oracle, $\delta=0.2$) & 38.38 & 15.23 & 30.88 \\
     
    \bottomrule
    \end{tabular}
    }

    \label{tab:wikilingua}
\end{table}

\section{Experiments}
\label{sec:experiment}

\subsection{Automatic Evaluation}
\label{sec:auto_experiment}

We use the F1 score of the standard ROUGE metric to evaluate all the models automatically. Results of Zh2EnSum and En2ZhSum  are shown in Table \ref{tab:ROUGE&BERTScore}.
For each comparable scenario where all the models have the same model size, the proposed CSC models achieve the best performance without any target length information in Zh-En$_{\rm norepeat}$. Specifically, our proposed CSC$_{\rm base}$ and CSC$_{\rm Multitask}$(SA) perform better than the multitask baselines with more parameters in Zh-En$_{\rm norepeat}$. Moreover, the CSC model with oracle $\gamma$ achieves even better results, suggesting that our model has a stronger performance for a more practical scenario where a desired summary length is given. 
Here, results of the Zh-En${\rm full}$ test set demonstrate that our model outperforms previous pipeline models.
Results of the Zh-En$_{\rm repeat}$ indicate that some models have overfitting issues, which will be further analyzed in Section~\ref{sec:overfitting}.
In En2ZhSum, CSC models get consistent improvement on R-2 and R-L metrics.


Meanwhile, we observe that the model CSC$_{\rm multitask}$(SA) outperforms the CSC model with $\delta=0.2$ and fixed compression rate. 
We speculate that it is because the model can be seen as trained with larger compression bins.
We will further discuss this phenomenon in Section~\ref{sec:bin}.
Another phenomenon appears that the performance of the CSC$_{\rm Multitask}$(SE) and the CSC$_{\rm Multitask}$(SD) fall behind their vanilla multitask versions. 
We attribute this to their large number of parameters, which makes them hard to train.

The results of NCLS+AG(SA) prove that only using our data augmentation method could bring improvements to the CLS models. Specifically, NCLS+AG(SA) surpassed the NCLS+MT(SA) in En2ZhSum dataset. It achieves the best result in Zh-En$_{\rm full}$, which is caused by its overfitting problem. We will analyze this in Section~\ref{sec:overfitting}.



We show the results of the WikiLingua dataset in Table \ref{tab:wikilingua}. 
The CSC model outperforms all the baseline models on R-1 and R-2, but there is an abnormal phenomenon that the fixed CSC model surpasses the oracle model. 
We inspect the results and find that the compression rates of some samples in WikiLingua are much larger than $100\%$ because they are erroneously constructed. 
With the oracle setting, our model will very likely process these samples as an MT task instead of a CLS task, leading to worse performance.

\subsection{Human Evaluation}
We conduct the human evaluation on the Zh2EnSum dataset. 
We pay three expert annotators who are proficient in English and Chinese to conduct the human evaluation on samples randomly selected from Zh-En$_{\rm norepeat}$. 
Informativeness, Fluency, and Conciseness are rated between 1 to 5 for each sample. 
We also evaluate the factual degree of samples, where each annotator is asked whether the summary is factual, resulting in a score ranging from 0 to 1. 

Specifically, each evaluation sample contains a source document, a target summary, and a system summary. Annotators are not told which system they are evaluating and asked to evaluate four aspects of the system summary. 1) Fluency: Maximum score is 5, one point is deducted when a syntax error appears. 2) Conciseness: Maximum score is 5, redundancy information including repetition, too long lengths, and extra information from the target summary will cause a score deduction. 3) Informativeness: Maximum score is 5, if the system summary loses information from the target summary or important information in the source target, its score will be deducted. 4) Factual Degree: if the system summary describes the consistent information to the source document, the score will be 1, otherwise 0. Three expert annotators are paid to evaluate 25 random samples in Zh-En$_{\rm norepeat}$.
Results are shown in Table~\ref{tab:human}. 

Overall, CSC models with augmented data achieve better results than baseline models.
Specifically, all the CSC$_{\rm multitask}$ models outperforms their corresponding NCLS+MT versions, proving the efficiency of the proposed augmented data. Moreover, CSC models outperform all the NCLS baselines in IF, FL, and CC metrics.


\begin{table}[t]
    \caption{Human evaluation results of Zh2EnSum dataset. IF, FL, CC, and FC indicate Informativeness, Fluency, Conciseness, and Factual Degree.}
  \centering
  \resizebox{0.80\linewidth}{!}{
    \begin{tabular}{lrrrr}
    \toprule
    Models & IF & FL & CC & FC \\
    \midrule
     NCLS+MT(SA) & 3.57 & 4.59 & 4.17 & 0.76  \\
     NCLS+MT(SE) & 3.56 & 4.69 & 4.12 & 0.63  \\
     NCLS+MT(SD) & 3.48 & 4.68 & 3.92 & 0.53   \\
     \midrule
     CSC$_{\rm Multitask}$(SA) & 3.72 & 4.67 & 4.24 & 0.64  \\
     CSC$_{\rm Multitask}$(SE) & \textbf{3.83}  & 4.79  & 4.16 & 0.76  \\
     CSC$_{\rm Multitask}$(SD) & 3.53  & \textbf{4.80} & 3.92 & 0.60 \\
     \midrule
     CSC$_{\rm base}$(fixed, $\delta = 0.2$) & 3.68 & 4.76 & \textbf{4.43} & 0.68 \\
     CSC$_{768}$(fixed, $\delta = 0.2$) & 3.69 & 4.79 & 4.28 & \textbf{0.77} \\
     
    \bottomrule
    \end{tabular}
    }

    \label{tab:human}
\end{table}

\subsection{Analysis on Overfitting}
\label{sec:overfitting}
Since Zh-En$_{\rm repeat}$ test set consists of samples appearing in the training set, its results can help us analyze the overfitting problem of each model. 
We find that, with better Zh-En$_{\rm norepeat}$ results, SA models and CSC$_{\rm base}$ get a worse Zh-En$_{\rm repeat}$ performance than other multitask models. 
This means that sharing all the parameters or possessing fewer parameters may relieve the model from overfitting. 
Also, only using the augmented data (NCLS+AG) leads to a severe overfitting problem. However, the introduction of the MT data can relieve it.

\subsection{Analysis on Controllability of $\gamma$}
\begin{figure}[t]
\centering
\includegraphics[width=0.99\columnwidth]{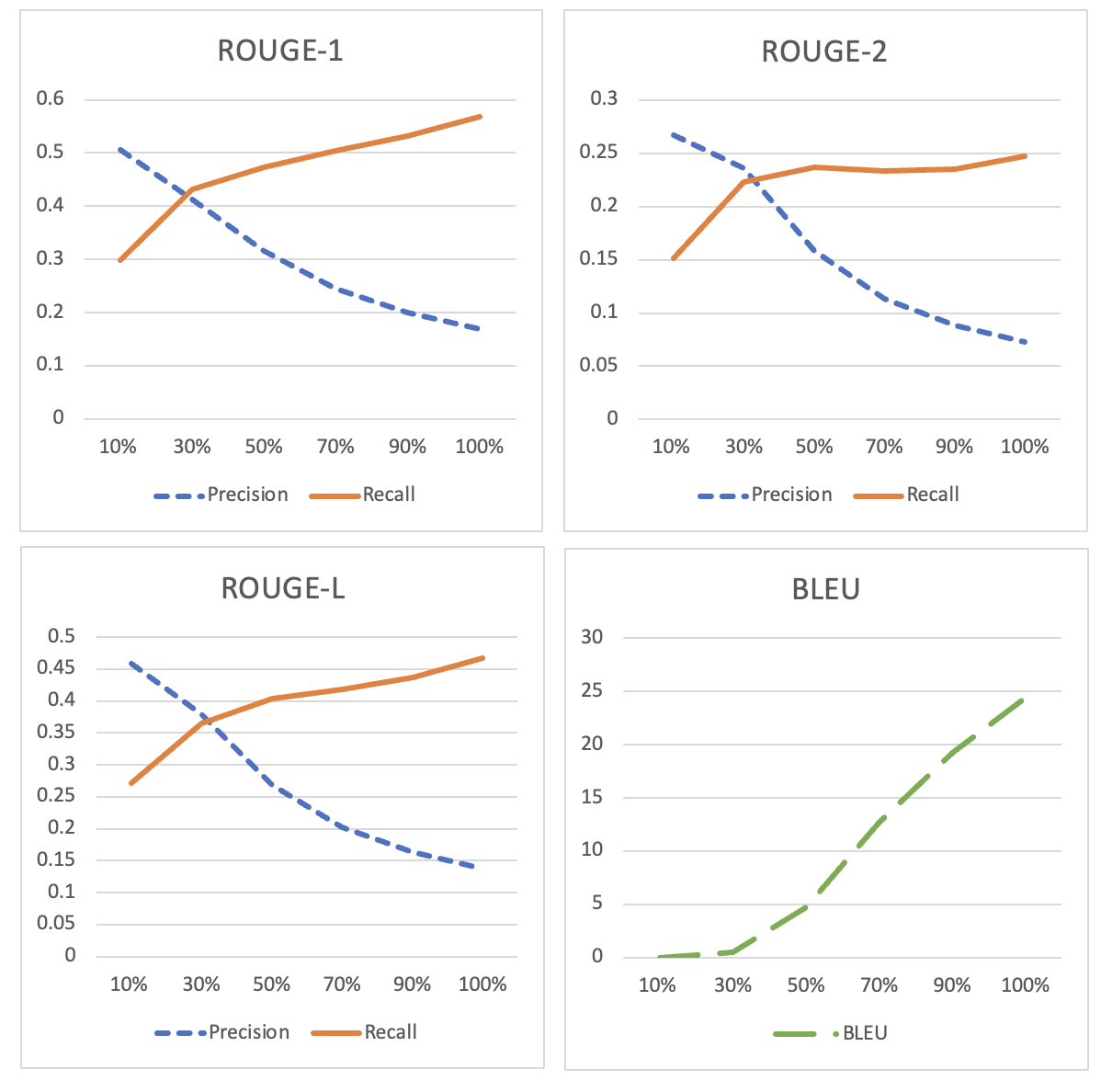} 
\caption{Results on the controllability of CSC$_{768}$~($\delta=0.2$) with different compression rates}
\label{fig:control}
\end{figure}

Apart from improving the performance of the model, a prominent advantage of the proposed CSC is that, it is able to generate summaries with different target lengths by adjusting the compression rate. 
To explore this feature, we generate summaries with different compression rate signals for each document in the test set. 
We show the recall and precision of the ROUGE metric in Figure~\ref{fig:control}. 
Obviously, as the compression rate raises, the ROUGE recall values also raises, which means that the model is absorbing more information into the final summary. 
However, since the expected information of the target summary is unchanged, the precision drops when the generated summary becomes longer. 

We also show the BLEU score with different compression rates where we apply the same model directly on the WMT17 Zh2En test set\footnote{We use sacreBLEU implementation~\citep{post-2018-call} for BLEU evaluation.}. 
It clearly shows the transition from CLS to MT: as the compression rate goes up, the BLEU score also raises until it completely meets the MT task. 


\begin{table}[t]
    \caption{Zh-En results of CSC models with different bin widths, where $0.033$ is the approximation of $\frac{1}{30}$.}
  \centering
  \resizebox{0.90\linewidth}{!}{
    \begin{tabular}{lrrrrrrr}
    \toprule
    CSC Models & \multicolumn{3}{c}{\textbf{Fixed $\gamma$}}  & \multicolumn{3}{c}{\textbf{Oracle $\gamma$}} & \multirow{2}*{{\makecell[r]{\textbf{Length}\\ \textbf{Variance} } }}
    \\
    \cmidrule(r{4pt}){2-4}  \cmidrule(r{4pt}){5-7} 
    ($\gamma = 0.25$) & R-1 & R-2 & R-L & R-1 & R-2 & R-L \\
    \midrule
     $\delta=0.020$ & 36.80 & 18.61 & 32.47 & 40.09 & 20.93 & 35.14 & \textbf{0.72} \\
     $\delta=0.033$ & 37.43 & 18.85 & 33.02  & \textbf{40.72} & 21.05 & 35.70 & 2.09 \\
     $\delta=0.050$ & 37.70 & 19.27 & 33.06 & 40.59 & 21.23 &\textbf{35.71} & 2.74 \\
     $\delta=0.200$ & \textbf{38.96} & \textbf{20.29} &\textbf{ 34.11} & 40.30 & \textbf{21.43} & 35.46 & 20.32  \\
    \bottomrule
    \end{tabular}
    }

    \label{tab:compression_bin}
\end{table}

\subsection{Analysis on Compression Rate Bin}
\label{sec:bin}
The bin width $\delta$ during discretization in CSC is an important hyper-parameter to control the precision. 
Numerically, a smaller $\delta$ allows the input compression rate to more precisely approximate the real length ratio of the document-summary pair. 
Hence, it is intuitive that a smaller $\delta$ should lead to a better oracle performance but a worse result given a fixed compression rate, and vice versa. 

To verify this hypothesis, we train various CSC$_{768}$ models with different bin widths and inference with the average compression rate of 25\% in the Zh2EnSum training set, shown in Table~\ref{tab:compression_bin}. 
We can observe that as $\delta$ becomes larger, the performance of the model gets better with a fixed $\gamma$. 
However, the performance of the oracle case does not drop as we expected, so we further measure the variance between the lengths of references and predictions, shown in the rightmost column in Table~\ref{tab:compression_bin}. 
Note that measuring length variance is almost equal to measuring the compression rate variance.
We can observe that, as $\delta$ becomes smaller, the variance also becomes smaller, which means the compression rate modeling does predict a more accurate length. 
We hypothesize that the slightly bad oracle performance is because the model extracts some wrong information.


The above observations motivate us to utilize the CSC training to meet different practical requirements. 
For example, in the application of news push notifications on different devices (e.g., iPhone and iPad), we need to strictly control the length of the generated summary, then we can use the model trained with small bins. 
If we only care about the overall quality of the summary rather than an expected length, we can set a relatively large bin. 



\begin{figure}[t]
\centering
\includegraphics[width=0.99\columnwidth]{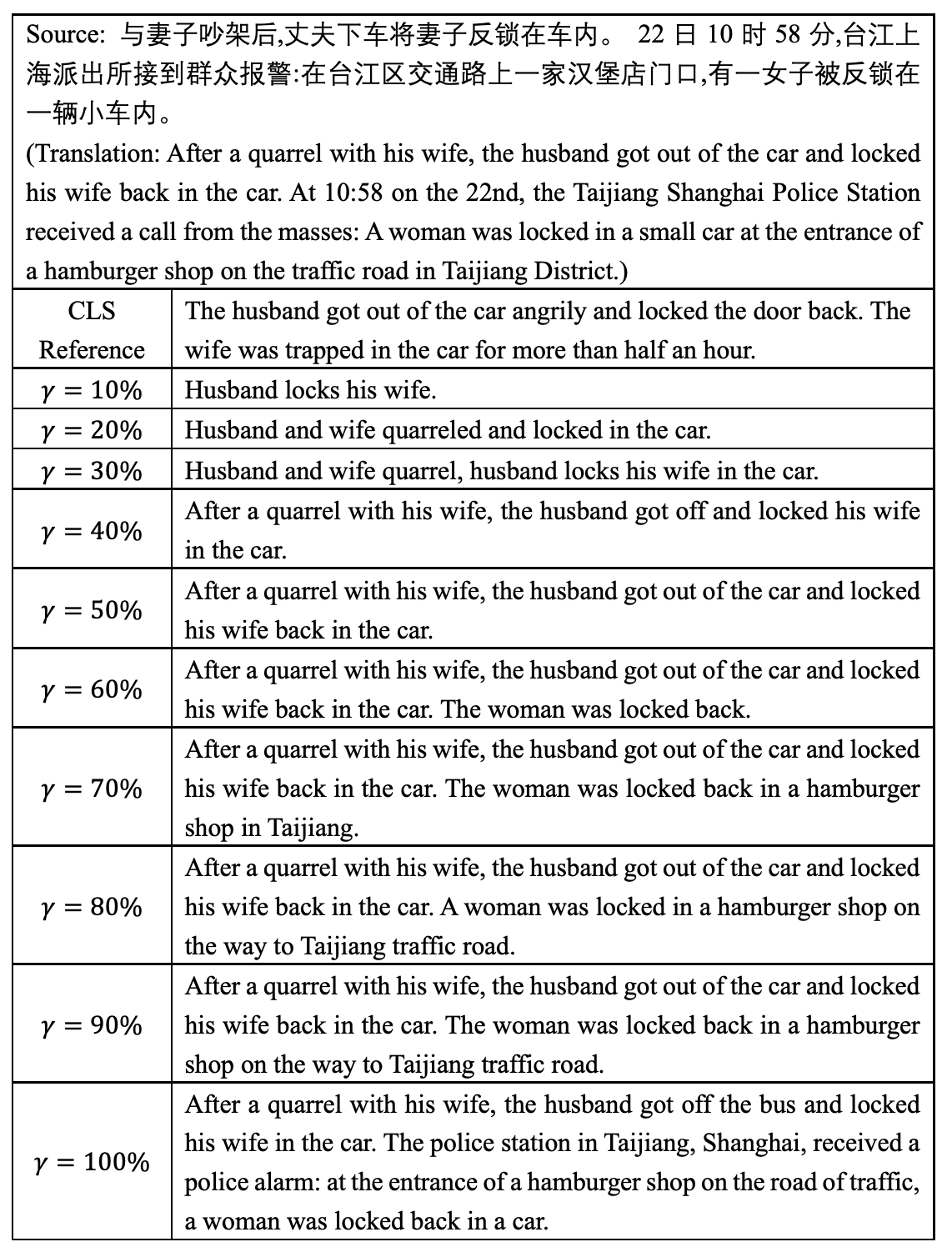} 
\caption{An example on the summaries generated with different compression rate. Note that here the $\gamma$ is computed with number of subword tokens.}
\label{fig:cr_example}
\end{figure}

\subsection{Case Study}
In this section, we show a case study on the controllability of our model, where results of different compression rates are shown. 
\begin{figure}[t]
\centering
\includegraphics[width=0.99\columnwidth]{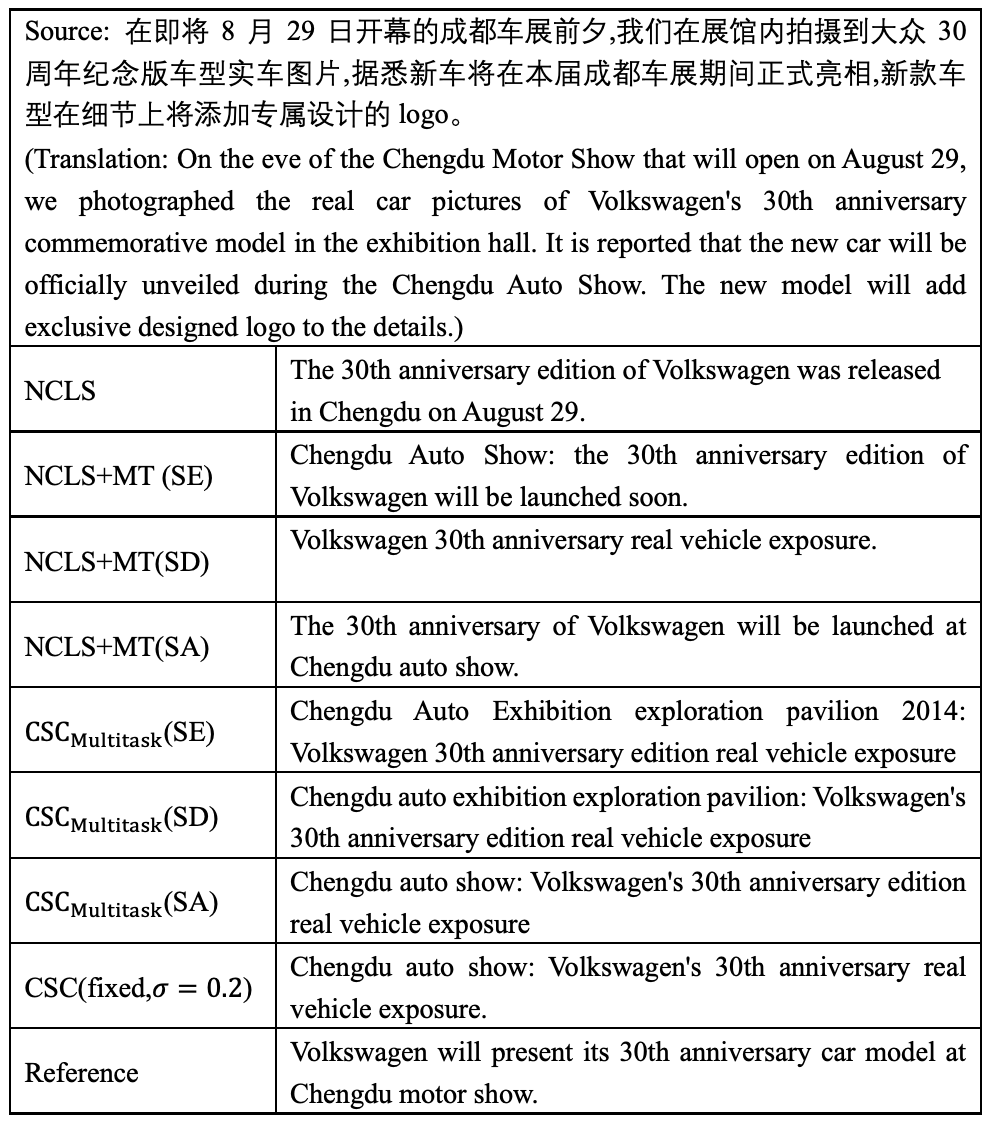} 
\caption{An example on the summaries generated by different systems. }
\label{fig:sample}
\end{figure}


\paragraph{Case study on controlability}
Guided by different compression rates, summary samples generated by our CSC$_{768}$ model~($\delta = 0.02$) are shown in Figure~\ref{fig:cr_example}. 
A clear upward trend of summary lengths is observed when $\gamma$ becomes larger. 
When $\gamma = 10\%$, the summary only contains four simple words `Husband locks his wife' which is fairly important for the article. 
As $\gamma$ becomes larger, the summary gets longer. 
Eventually, the generated text becomes the translation of the source text.
This case shows the practical controllability of the proposed CSC model.
However, the model still has its weakness. 
For example, when $\gamma=100\%$, a translation mistake is made by the model.
\begin{CJK*}{UTF8}{gbsn}
The word `车' is translated into `bus' instead of `car'.
\end{CJK*} 
The model also ignores the time information `at 10:58 on the 22nd'.
This shows that the summarization training may have a negative effect on the MT task.

\paragraph{Case study on quality}
For summary quality, we show system summaries generated by different models in Figure~\ref{fig:sample}. 
The output of baseline models lost some important information such as `Chengdu Auto Show'. 
Some baselines such as NCLS+MT(SD) suffer from fluency issues. Specifically, non-factual outputs are generated by some baselines such as `was launched', indicating the existence of a hallucination problem. 
In this case, the oracle $\gamma$ and the fixed $\gamma$ are in the same compression bin. 
Hence, we omit the oracle result. Overall, the summary produced by CSC is the most similar to the reference summary.

\section{Discussion}
We claim that our proposed CSC task brings a wider image for the application of cross-lingual summarization. 
With the development of technology, summary requirements for different lengths become more mightiness. 
For example, one news document may need different versions of summaries: watch/phone notifications, twitter short texts, search result summaries, web page summaries, etc. 
Our method provides a simple yet effective method to achieve diversity. 
With one CSC model, summaries with various lengths can be generated at once (through parallel computing), fulfilling different requirements.  
In general, our method is practical in various realistic scenarios.

Besides the MT task, another high-resource task related to CLS is the Monolingual Summarization task. 
We can readily incorporate the MS task into the CSC framework. 
In our preliminary experiments, we found that generating summaries in its own language may distract the performance of the cross-lingual summary, leading to worse performance than our proposed approach. 
However, we think that incorporating the MS task should benefit the CLS task and need more exploration. 
We will leave it for future work.

\section{Conclusion}
In this paper, we propose a novel task, Cross-lingual Summarization with Compression Rate. 
We propose a simple yet effective data-augment method and a modified Transformer model to unify the Machine Translation and Cross-lingual Summarization tasks. 
Experiments demonstrate that our approach not only improves the performance of the CLS task,
but also provides a promising direction for the controllability of summaries in desired lengths. 
We believe it will have great potentials in real application scenarios.

\section*{Acknowledgements}
This work is supported by Joint Funds of the National Natural Science
Foundation of China (Grant No. U19B2020).

We appreciate the helpful discussions with Sanxing Chen, Xiaochen Liu, Jiawei Li, Weilun Wu and Yizhe Yang. We also thank all the anonymous reviewers for their insightful suggestions.

\bibliographystyle{ACM-Reference-Format}
\bibliography{sample-base}


\begin{thebibliography}{31}


\ifx \showCODEN    \undefined \def \showCODEN     #1{\unskip}     \fi
\ifx \showDOI      \undefined \def \showDOI       #1{#1}\fi
\ifx \showISBNx    \undefined \def \showISBNx     #1{\unskip}     \fi
\ifx \showISBNxiii \undefined \def \showISBNxiii  #1{\unskip}     \fi
\ifx \showISSN     \undefined \def \showISSN      #1{\unskip}     \fi
\ifx \showLCCN     \undefined \def \showLCCN      #1{\unskip}     \fi
\ifx \shownote     \undefined \def \shownote      #1{#1}          \fi
\ifx \showarticletitle \undefined \def \showarticletitle #1{#1}   \fi
\ifx \showURL      \undefined \def \showURL       {\relax}        \fi
\providecommand\bibfield[2]{#2}
\providecommand\bibinfo[2]{#2}
\providecommand\natexlab[1]{#1}
\providecommand\showeprint[2][]{arXiv:#2}

\bibitem[Babar and Patil(2015)]%
        {babar2015improving}
\bibfield{author}{\bibinfo{person}{SA Babar} {and} \bibinfo{person}{Pallavi~D
  Patil}.} \bibinfo{year}{2015}\natexlab{}.
\newblock \showarticletitle{Improving performance of text summarization}.
\newblock \bibinfo{journal}{\emph{Procedia Computer Science}}
  \bibinfo{volume}{46} (\bibinfo{year}{2015}), \bibinfo{pages}{354--363}.
\newblock


\bibitem[Bai et~al\mbox{.}(2021)]%
        {bai-etal-2021-cross}
\bibfield{author}{\bibinfo{person}{Yu Bai}, \bibinfo{person}{Yang Gao}, {and}
  \bibinfo{person}{Heyan Huang}.} \bibinfo{year}{2021}\natexlab{}.
\newblock \showarticletitle{Cross-Lingual Abstractive Summarization with
  Limited Parallel Resources}. In \bibinfo{booktitle}{\emph{Proceedings of the
  59th Annual Meeting of the Association for Computational Linguistics and the
  11th International Joint Conference on Natural Language Processing (Volume 1:
  Long Papers)}}. \bibinfo{publisher}{Association for Computational
  Linguistics}, \bibinfo{address}{Online}, \bibinfo{pages}{6910--6924}.
\newblock
\urldef\tempurl%
\url{https://doi.org/10.18653/v1/2021.acl-long.538}
\showDOI{\tempurl}


\bibitem[Cao et~al\mbox{.}(2020)]%
        {cao2020jointly}
\bibfield{author}{\bibinfo{person}{Yue Cao}, \bibinfo{person}{Hui Liu}, {and}
  \bibinfo{person}{Xiaojun Wan}.} \bibinfo{year}{2020}\natexlab{}.
\newblock \showarticletitle{Jointly Learning to Align and Summarize for Neural
  Cross-Lingual Summarization}. In \bibinfo{booktitle}{\emph{Proceedings of the
  58th Annual Meeting of the Association for Computational Linguistics}}.
  \bibinfo{pages}{6220--6231}.
\newblock


\bibitem[Fan et~al\mbox{.}(2018)]%
        {fan2018controllable}
\bibfield{author}{\bibinfo{person}{Angela Fan}, \bibinfo{person}{David
  Grangier}, {and} \bibinfo{person}{Michael Auli}.}
  \bibinfo{year}{2018}\natexlab{}.
\newblock \showarticletitle{Controllable Abstractive Summarization}. In
  \bibinfo{booktitle}{\emph{Proceedings of the 2nd Workshop on Neural Machine
  Translation and Generation}}. \bibinfo{pages}{45--54}.
\newblock


\bibitem[Hahn and Mani(2000)]%
        {hahn2000challenges}
\bibfield{author}{\bibinfo{person}{Udo Hahn} {and} \bibinfo{person}{Inderjeet
  Mani}.} \bibinfo{year}{2000}\natexlab{}.
\newblock \showarticletitle{The challenges of automatic summarization}.
\newblock \bibinfo{journal}{\emph{Computer}} \bibinfo{volume}{33},
  \bibinfo{number}{11} (\bibinfo{year}{2000}), \bibinfo{pages}{29--36}.
\newblock


\bibitem[Hu et~al\mbox{.}(2015)]%
        {hu2015lcsts}
\bibfield{author}{\bibinfo{person}{Baotian Hu}, \bibinfo{person}{Qingcai Chen},
  {and} \bibinfo{person}{Fangze Zhu}.} \bibinfo{year}{2015}\natexlab{}.
\newblock \showarticletitle{LCSTS: A Large Scale Chinese Short Text
  Summarization Dataset}. In \bibinfo{booktitle}{\emph{Proceedings of the 2015
  Conference on Empirical Methods in Natural Language Processing}}.
  \bibinfo{pages}{1967--1972}.
\newblock


\bibitem[Kikuchi et~al\mbox{.}(2016)]%
        {kikuchi2016controlling}
\bibfield{author}{\bibinfo{person}{Yuta Kikuchi}, \bibinfo{person}{Graham
  Neubig}, \bibinfo{person}{Ryohei Sasano}, \bibinfo{person}{Hiroya Takamura},
  {and} \bibinfo{person}{Manabu Okumura}.} \bibinfo{year}{2016}\natexlab{}.
\newblock \showarticletitle{Controlling Output Length in Neural
  Encoder-Decoders}. In \bibinfo{booktitle}{\emph{Proceedings of the 2016
  Conference on Empirical Methods in Natural Language Processing}}.
  \bibinfo{pages}{1328--1338}.
\newblock


\bibitem[Kudo and Richardson(2018)]%
        {kudo-richardson-2018-sentencepiece}
\bibfield{author}{\bibinfo{person}{Taku Kudo} {and} \bibinfo{person}{John
  Richardson}.} \bibinfo{year}{2018}\natexlab{}.
\newblock \showarticletitle{{S}entence{P}iece: A simple and language
  independent subword tokenizer and detokenizer for Neural Text Processing}. In
  \bibinfo{booktitle}{\emph{Proceedings of the 2018 Conference on Empirical
  Methods in Natural Language Processing: System Demonstrations}}.
  \bibinfo{publisher}{Association for Computational Linguistics},
  \bibinfo{address}{Brussels, Belgium}, \bibinfo{pages}{66--71}.
\newblock
\urldef\tempurl%
\url{https://doi.org/10.18653/v1/D18-2012}
\showDOI{\tempurl}


\bibitem[Ladhak et~al\mbox{.}(2020)]%
        {ladhak2020wikilingua}
\bibfield{author}{\bibinfo{person}{Faisal Ladhak}, \bibinfo{person}{Esin
  Durmus}, \bibinfo{person}{Claire Cardie}, {and} \bibinfo{person}{Kathleen
  McKeown}.} \bibinfo{year}{2020}\natexlab{}.
\newblock \showarticletitle{WikiLingua: A New Benchmark Dataset for
  Multilingual Abstractive Summarization}. In
  \bibinfo{booktitle}{\emph{Proceedings of the 2020 Conference on Empirical
  Methods in Natural Language Processing: Findings}}.
  \bibinfo{pages}{4034--4048}.
\newblock


\bibitem[Lin(2004)]%
        {lin2004rouge}
\bibfield{author}{\bibinfo{person}{Chin-Yew Lin}.}
  \bibinfo{year}{2004}\natexlab{}.
\newblock \showarticletitle{Rouge: A package for automatic evaluation of
  summaries}.
\newblock \bibinfo{journal}{\emph{Text Summarization Branches Out}}
  (\bibinfo{year}{2004}).
\newblock


\bibitem[Liu et~al\mbox{.}(2020)]%
        {liu2020multilingual}
\bibfield{author}{\bibinfo{person}{Yinhan Liu}, \bibinfo{person}{Jiatao Gu},
  \bibinfo{person}{Naman Goyal}, \bibinfo{person}{Xian Li},
  \bibinfo{person}{Sergey Edunov}, \bibinfo{person}{Marjan Ghazvininejad},
  \bibinfo{person}{Mike Lewis}, {and} \bibinfo{person}{Luke Zettlemoyer}.}
  \bibinfo{year}{2020}\natexlab{}.
\newblock \showarticletitle{Multilingual denoising pre-training for neural
  machine translation}.
\newblock \bibinfo{journal}{\emph{Transactions of the Association for
  Computational Linguistics}}  \bibinfo{volume}{8} (\bibinfo{year}{2020}),
  \bibinfo{pages}{726--742}.
\newblock


\bibitem[Liu et~al\mbox{.}(2018)]%
        {liu2018controlling}
\bibfield{author}{\bibinfo{person}{Yizhu Liu}, \bibinfo{person}{Zhiyi Luo},
  {and} \bibinfo{person}{Kenny Zhu}.} \bibinfo{year}{2018}\natexlab{}.
\newblock \showarticletitle{Controlling length in abstractive summarization
  using a convolutional neural network}. In
  \bibinfo{booktitle}{\emph{Proceedings of the 2018 Conference on Empirical
  Methods in Natural Language Processing}}. \bibinfo{pages}{4110--4119}.
\newblock


\bibitem[Luong et~al\mbox{.}(2015)]%
        {luong2015multi}
\bibfield{author}{\bibinfo{person}{Minh-Thang Luong}, \bibinfo{person}{Quoc~V
  Le}, \bibinfo{person}{Ilya Sutskever}, \bibinfo{person}{Oriol Vinyals}, {and}
  \bibinfo{person}{Lukasz Kaiser}.} \bibinfo{year}{2015}\natexlab{}.
\newblock \showarticletitle{Multi-task sequence to sequence learning}.
\newblock \bibinfo{journal}{\emph{arXiv preprint arXiv:1511.06114}}
  (\bibinfo{year}{2015}).
\newblock


\bibitem[Makino et~al\mbox{.}(2019)]%
        {makino2019global}
\bibfield{author}{\bibinfo{person}{Takuya Makino}, \bibinfo{person}{Tomoya
  Iwakura}, \bibinfo{person}{Hiroya Takamura}, {and} \bibinfo{person}{Manabu
  Okumura}.} \bibinfo{year}{2019}\natexlab{}.
\newblock \showarticletitle{Global optimization under length constraint for
  neural text summarization}. In \bibinfo{booktitle}{\emph{Proceedings of the
  57th Annual Meeting of the Association for Computational Linguistics}}.
  \bibinfo{pages}{1039--1048}.
\newblock


\bibitem[Neto et~al\mbox{.}(2002)]%
        {neto2002automatic}
\bibfield{author}{\bibinfo{person}{Joel~Larocca Neto}, \bibinfo{person}{Alex~A
  Freitas}, {and} \bibinfo{person}{Celso~AA Kaestner}.}
  \bibinfo{year}{2002}\natexlab{}.
\newblock \showarticletitle{Automatic text summarization using a machine
  learning approach}. In \bibinfo{booktitle}{\emph{Brazilian symposium on
  artificial intelligence}}. Springer, \bibinfo{pages}{205--215}.
\newblock


\bibitem[Nomoto and Matsumoto(2001)]%
        {nomoto2001new}
\bibfield{author}{\bibinfo{person}{Tadashi Nomoto} {and} \bibinfo{person}{Yuji
  Matsumoto}.} \bibinfo{year}{2001}\natexlab{}.
\newblock \showarticletitle{A new approach to unsupervised text summarization}.
  In \bibinfo{booktitle}{\emph{Proceedings of the 24th annual international ACM
  SIGIR conference on Research and development in information retrieval}}.
  \bibinfo{pages}{26--34}.
\newblock


\bibitem[Ott et~al\mbox{.}(2019)]%
        {ott2019fairseq}
\bibfield{author}{\bibinfo{person}{Myle Ott}, \bibinfo{person}{Sergey Edunov},
  \bibinfo{person}{Alexei Baevski}, \bibinfo{person}{Angela Fan},
  \bibinfo{person}{Sam Gross}, \bibinfo{person}{Nathan Ng},
  \bibinfo{person}{David Grangier}, {and} \bibinfo{person}{Michael Auli}.}
  \bibinfo{year}{2019}\natexlab{}.
\newblock \showarticletitle{fairseq: A Fast, Extensible Toolkit for Sequence
  Modeling}. In \bibinfo{booktitle}{\emph{Proceedings of NAACL-HLT 2019:
  Demonstrations}}.
\newblock


\bibitem[Ouyang et~al\mbox{.}(2019)]%
        {ouyang2019robust}
\bibfield{author}{\bibinfo{person}{Jessica Ouyang}, \bibinfo{person}{Boya
  Song}, {and} \bibinfo{person}{Kathleen McKeown}.}
  \bibinfo{year}{2019}\natexlab{}.
\newblock \showarticletitle{A robust abstractive system for cross-lingual
  summarization}. In \bibinfo{booktitle}{\emph{Proceedings of the 2019
  Conference of the North American Chapter of the Association for Computational
  Linguistics: Human Language Technologies, Volume 1 (Long and Short Papers)}}.
  \bibinfo{pages}{2025--2031}.
\newblock


\bibitem[Perez-Beltrachini and Lapata(2021)]%
        {perez-beltrachini-lapata-2021-models}
\bibfield{author}{\bibinfo{person}{Laura Perez-Beltrachini} {and}
  \bibinfo{person}{Mirella Lapata}.} \bibinfo{year}{2021}\natexlab{}.
\newblock \showarticletitle{Models and Datasets for Cross-Lingual
  Summarisation}. In \bibinfo{booktitle}{\emph{Proceedings of the 2021
  Conference on Empirical Methods in Natural Language Processing}}.
  \bibinfo{publisher}{Association for Computational Linguistics},
  \bibinfo{address}{Online and Punta Cana, Dominican Republic},
  \bibinfo{pages}{9408--9423}.
\newblock
\urldef\tempurl%
\url{https://aclanthology.org/2021.emnlp-main.742}
\showURL{%
\tempurl}


\bibitem[Post(2018)]%
        {post-2018-call}
\bibfield{author}{\bibinfo{person}{Matt Post}.}
  \bibinfo{year}{2018}\natexlab{}.
\newblock \showarticletitle{A Call for Clarity in Reporting {BLEU} Scores}. In
  \bibinfo{booktitle}{\emph{Proceedings of the Third Conference on Machine
  Translation: Research Papers}}. \bibinfo{publisher}{Association for
  Computational Linguistics}, \bibinfo{address}{Belgium, Brussels},
  \bibinfo{pages}{186--191}.
\newblock
\urldef\tempurl%
\url{https://www.aclweb.org/anthology/W18-6319}
\showURL{%
\tempurl}


\bibitem[Takase and Okazaki(2019)]%
        {takase2019positional}
\bibfield{author}{\bibinfo{person}{Sho Takase} {and} \bibinfo{person}{Naoaki
  Okazaki}.} \bibinfo{year}{2019}\natexlab{}.
\newblock \showarticletitle{Positional Encoding to Control Output Sequence
  Length}. In \bibinfo{booktitle}{\emph{Proceedings of the 2019 Conference of
  the North American Chapter of the Association for Computational Linguistics:
  Human Language Technologies, Volume 1 (Long and Short Papers)}}.
  \bibinfo{pages}{3999--4004}.
\newblock


\bibitem[Takase and Okazaki(2020)]%
        {takase2020multi}
\bibfield{author}{\bibinfo{person}{Sho Takase} {and} \bibinfo{person}{Naoaki
  Okazaki}.} \bibinfo{year}{2020}\natexlab{}.
\newblock \showarticletitle{Multi-Task Learning for Cross-Lingual Abstractive
  Summarization}.
\newblock \bibinfo{journal}{\emph{arXiv preprint arXiv:2010.07503}}
  (\bibinfo{year}{2020}).
\newblock


\bibitem[Tas and Kiyani(2007)]%
        {tas2007survey}
\bibfield{author}{\bibinfo{person}{Oguzhan Tas} {and} \bibinfo{person}{Farzad
  Kiyani}.} \bibinfo{year}{2007}\natexlab{}.
\newblock \showarticletitle{A survey automatic text summarization}.
\newblock \bibinfo{journal}{\emph{PressAcademia Procedia}} \bibinfo{volume}{5},
  \bibinfo{number}{1} (\bibinfo{year}{2007}), \bibinfo{pages}{205--213}.
\newblock


\bibitem[Vaswani et~al\mbox{.}(2017)]%
        {vaswani2017attention}
\bibfield{author}{\bibinfo{person}{Ashish Vaswani}, \bibinfo{person}{Noam
  Shazeer}, \bibinfo{person}{Niki Parmar}, \bibinfo{person}{Jakob Uszkoreit},
  \bibinfo{person}{Llion Jones}, \bibinfo{person}{Aidan~N Gomez},
  \bibinfo{person}{{\L}ukasz Kaiser}, {and} \bibinfo{person}{Illia
  Polosukhin}.} \bibinfo{year}{2017}\natexlab{}.
\newblock \showarticletitle{Attention is all you need}. In
  \bibinfo{booktitle}{\emph{Advances in neural information processing
  systems}}. \bibinfo{pages}{5998--6008}.
\newblock


\bibitem[Wan(2011)]%
        {wan2011using}
\bibfield{author}{\bibinfo{person}{Xiaojun Wan}.}
  \bibinfo{year}{2011}\natexlab{}.
\newblock \showarticletitle{Using bilingual information for cross-language
  document summarization}. In \bibinfo{booktitle}{\emph{Proceedings of the 49th
  Annual Meeting of the Association for Computational Linguistics: Human
  Language Technologies}}. \bibinfo{pages}{1546--1555}.
\newblock


\bibitem[Wan et~al\mbox{.}(2010)]%
        {wan2010cross}
\bibfield{author}{\bibinfo{person}{Xiaojun Wan}, \bibinfo{person}{Huiying Li},
  {and} \bibinfo{person}{Jianguo Xiao}.} \bibinfo{year}{2010}\natexlab{}.
\newblock \showarticletitle{Cross-language document summarization based on
  machine translation quality prediction}. In
  \bibinfo{booktitle}{\emph{Proceedings of the 48th Annual Meeting of the
  Association for Computational Linguistics}}. \bibinfo{pages}{917--926}.
\newblock


\bibitem[Xu et~al\mbox{.}(2020)]%
        {xu2020mixed}
\bibfield{author}{\bibinfo{person}{Ruochen Xu}, \bibinfo{person}{Chenguang
  Zhu}, \bibinfo{person}{Yu Shi}, \bibinfo{person}{Michael Zeng}, {and}
  \bibinfo{person}{Xuedong Huang}.} \bibinfo{year}{2020}\natexlab{}.
\newblock \showarticletitle{Mixed-Lingual Pre-training for Cross-lingual
  Summarization}. In \bibinfo{booktitle}{\emph{Proceedings of the 1st
  Conference of the Asia-Pacific Chapter of the Association for Computational
  Linguistics and the 10th International Joint Conference on Natural Language
  Processing}}. \bibinfo{pages}{536--541}.
\newblock


\bibitem[Yeh et~al\mbox{.}(2005)]%
        {yeh2005text}
\bibfield{author}{\bibinfo{person}{Jen-Yuan Yeh}, \bibinfo{person}{Hao-Ren Ke},
  \bibinfo{person}{Wei-Pang Yang}, {and} \bibinfo{person}{I-Heng Meng}.}
  \bibinfo{year}{2005}\natexlab{}.
\newblock \showarticletitle{Text summarization using a trainable summarizer and
  latent semantic analysis}.
\newblock \bibinfo{journal}{\emph{Information processing \& management}}
  \bibinfo{volume}{41}, \bibinfo{number}{1} (\bibinfo{year}{2005}),
  \bibinfo{pages}{75--95}.
\newblock


\bibitem[Zhang et~al\mbox{.}(2016)]%
        {zhang2016abstractive}
\bibfield{author}{\bibinfo{person}{Jiajun Zhang}, \bibinfo{person}{Yu Zhou},
  {and} \bibinfo{person}{Chengqing Zong}.} \bibinfo{year}{2016}\natexlab{}.
\newblock \showarticletitle{Abstractive cross-language summarization via
  translation model enhanced predicate argument structure fusing}.
\newblock \bibinfo{journal}{\emph{IEEE/ACM Transactions on Audio, Speech, and
  Language Processing}} \bibinfo{volume}{24}, \bibinfo{number}{10}
  (\bibinfo{year}{2016}), \bibinfo{pages}{1842--1853}.
\newblock


\bibitem[Zhu et~al\mbox{.}(2019)]%
        {zhu2019ncls}
\bibfield{author}{\bibinfo{person}{Junnan Zhu}, \bibinfo{person}{Qian Wang},
  \bibinfo{person}{Yining Wang}, \bibinfo{person}{Yu Zhou},
  \bibinfo{person}{Jiajun Zhang}, \bibinfo{person}{Shaonan Wang}, {and}
  \bibinfo{person}{Chengqing Zong}.} \bibinfo{year}{2019}\natexlab{}.
\newblock \showarticletitle{NCLS: Neural Cross-Lingual Summarization}. In
  \bibinfo{booktitle}{\emph{Proceedings of the 2019 Conference on Empirical
  Methods in Natural Language Processing and the 9th International Joint
  Conference on Natural Language Processing (EMNLP-IJCNLP)}}.
  \bibinfo{pages}{3045--3055}.
\newblock


\bibitem[Zhu et~al\mbox{.}(2020)]%
        {zhu2020attend}
\bibfield{author}{\bibinfo{person}{Junnan Zhu}, \bibinfo{person}{Yu Zhou},
  \bibinfo{person}{Jiajun Zhang}, {and} \bibinfo{person}{Chengqing Zong}.}
  \bibinfo{year}{2020}\natexlab{}.
\newblock \showarticletitle{Attend, translate and summarize: An efficient
  method for neural cross-lingual summarization}. In
  \bibinfo{booktitle}{\emph{Proceedings of the 58th Annual Meeting of the
  Association for Computational Linguistics}}. \bibinfo{pages}{1309--1321}.
\newblock


\end{thebibliography}










\end{document}